\pdfoutput=1

\documentclass[11pt]{article}
\usepackage{graphicx}
\usepackage{multirow}
\usepackage{float}
\usepackage{afterpage}
\usepackage{placeins}
\usepackage{booktabs}
\usepackage{graphicx,lscape,rotating}

\usepackage[preprint]{acl}

\usepackage{times}
\usepackage{latexsym}

\usepackage[T1]{fontenc}

\usepackage[utf8]{inputenc}

\usepackage{microtype}

\usepackage{inconsolata}

\title{LoraMap: Harnessing the Power of LoRA Connections}

\author{Hyeryun Park$^{1,2}$, Jeongwon Kwak$^{1,2}$, Dongsuk Jang$^{1,2}$, Sumin Park$^{4}$, Jinwook Choi$^{2,3,4}$ \\
        $^{1}$Interdisciplinary Program for Bioengineering, Graduate School, Seoul National University \\
        $^{2}$Integrated Major in Innovative Medical Science, Graduate School, Seoul National University \\
        $^{3}$Department of Biomedical Engineering, College of Medicine, Seoul National University\\
        $^{4}$Medical Research Center, Institute of Medical and Biological Engineering, \\ Seoul National University \\}

\usepackage{amsmath,amssymb}

\begin{document}
\maketitle
\begin{abstract}
Fact-checking techniques can mitigate hallucinations in Large Language Models (LLMs), a prominent issue in specialized domains. As parameter-efficient techniques such as Low-Rank Adaptation (LoRA) can overcome substantial computational overhead, some studies have explored the integration of multiple LoRAs. While previous studies focus on parallel integration, this paper investigates methods to establish connections among multiple LoRAs. We create three reasoning datasets tailored to fact-checking and fine-tune individual LoRAs, allowing them to view and reason from diverse perspectives. Then, we explore strategies for allocating these reasoning LoRAs and introduce LoraMap, an approach to map connections between them. The results of the fact-checking task demonstrate that the performance of LoraMap is superior to LoraHub, an existing method for integrating LoRAs. LoraMap also outperforms with significantly fewer trainable parameters than LoraConcat, which concatenates LoRAs and further fine-tunes them.
\end{abstract}

\section{Introduction}

With the rapid progress in research leveraging Large Language Models (LLMs) such as GPT-4 \citep{Touvron2023}, PaLM \citep{Chowdhery2023}, LLaMA \citep{Touvron2023_2}, and Flan-T5 \citep{Chung2022} in various natural language processing tasks, several challenges have also emerged. The model can pose a significant risk to reliability and trustworthiness due to the issue of generating false information, known as hallucination \citep{Ji2023}. One way to alleviate this problem is using fact-checking to verify LLM outputs or given claims \citep{Gupta2022, Chamoun2023}.

\begin{figure*}[h!]
    \centering
    \includegraphics[scale=0.55]{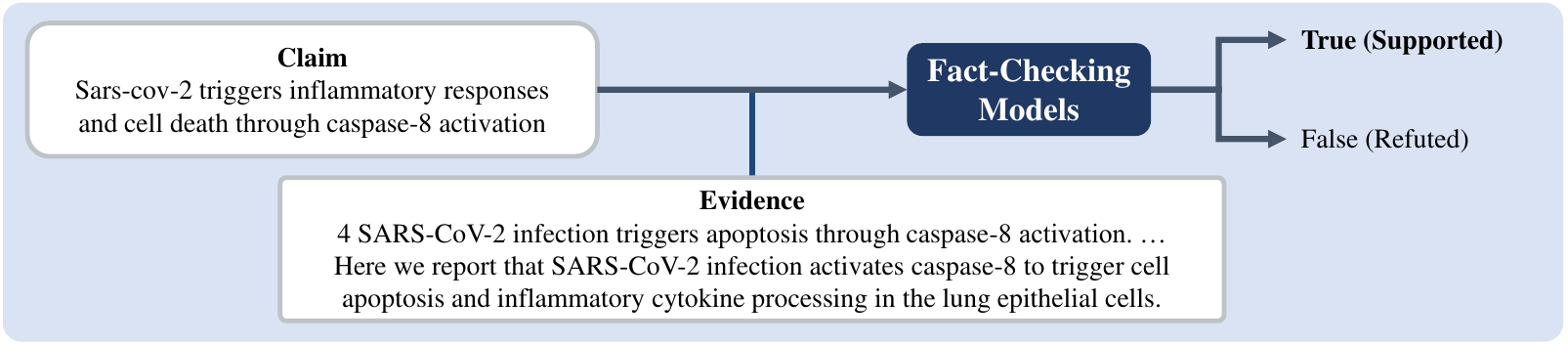}
    \caption{A fact-checking task classifies a claim into true or false based on the corresponding evidence.}
    \label{fact-checking}
\end{figure*}

As in Figure~\ref{fact-checking}, a fact-checking process classifies a claim into true, false, or more sophisticated labels based on textual evidence such as Wikipedia passages, news articles, and other relevant documents \citep{Thorne2018,Guo2022}. 
In biomedical and health domains, serious problems can arise when people perceive false information as truth, highlighting the importance of fact-checking.
Accordingly, many studies have been explored, resulting in the development of datasets: SciFact \citep{Wadden2020}, PubHealth \citep{KotonyaToni2020}, COVID-Fact \citep{Saakyan2021}, and HealthVer \citep{Sarrouti2021}. This paper focuses on small datasets, COVID-Fact and SciFact.

Another challenge is that fine-tuning the LLMs requires high computational demands. Parameter-efficient fine-tuning techniques can address this issue, especially Low-rank adaptations (LoRA) \citep{Hu2021}.
As numerous task-specific LoRAs have appeared, some studies have explored the integration of these modules to serve auxiliary roles in addressing new tasks \citep{Huang2023,Liu2023Moelora,Gao2023,Li2024mixlora,Dou2023loramoe}.
Among these methods, LoraHub \citep{Huang2023} learns weights for each LoRA and computes their weighted sum in parallel, which may weaken the influence of the pivotal LoRA.

This paper investigates methods for establishing connections among LoRAs to exchange their specialized insights as an alternative to parallel integration. Our main contributions are as follows:
\begin{itemize}\setlength{\itemsep}{0pt}
    \item We create three reasoning datasets tailored to fact-checking and fine-tune LoRA for each dataset, allowing them to infer from various perspectives.
    \item We investigate how to integrate these reasoning LoRAs and introduce LoraMap. Inspired by the information-processing behavior of the human brain in neuroscience, it learns connections rather than a linear sum of LoRAs.
    \item The results on the COVID-Fact and SciFact datasets demonstrate that LoraMap exhibits superior performance than LoraHub and also outperforms LoraConcat even with significantly fewer trainable parameters.
\end{itemize}

\section{Related Work}
\subsection{Fact-checking using LLMs}
Recent studies have explored the potential of LLMs for fact-checking through zero-shot prompting \citep{Chern2023factool,Li2023selfchecker,Wang2023factcheck}, hierarchical step-by-step prompting with external knowledge \citep{zhang2023towards}, multi-agent debating \citep{du2023improving}, question answering in intermediate steps \citep{pan2023qacheck}, and evaluating factuality within lengthy text \citep{min2023factscore}. 
These approaches employ advanced prompting strategies and external knowledge to enhance reasoning.
The results demonstrate the effectiveness of leveraging LLMs for fact-checking and suggest further advancements in factual reasoning \citep{laban2023,wang2023survey}.

\subsection{Biomedical Fact-checking}
The rapid expansion of evidence, encompassing biomedical articles and literature, has made manual fact-checking challenging and time-consuming.
Several studies have attempted to construct biomedical fact-checking datasets and train diverse models.
For the PubHealth dataset, the SciBERT model achieves the highest f1-score among the BERT models \citep{KotonyaToni2020}.
On the SciFact leaderboard\footnote{\url{https://leaderboard.allenai.org/scifact/submissions/public}}, the best model is MultiVerS \citep{wadden2022multivers}, a Longformer model \citep{Beltagy2020} that is trained with rationale sentence selection and fact-checking label prediction.
For COVID-Fact, the RoBERTa model is fine-tuned on fact-checking and entailment inference datasets \citep{Saakyan2021}.
In the case of HealthVer, the T5-base model performed better than the BERT models \citep{Sarrouti2021}.
The COVID-Fact and SciFact are small datasets containing fewer than 5,000 claim-evidence pairs, whereas the HealthVer and PubHealth datasets include more than 10,000 instances.

\subsection{Parameter-efficient Fine-tuning}
Several studies have introduced parameter-efficient fine-tuning techniques that freeze the original model parameters and only fine-tune a few additional parameters.
Adapter tuning \citep{Houlsby2019,Pfeiffer2020} inserts a layer into each transformer layer, which consists of a down-projection feed-forward layer, a non-linearity function, and an up-projection feed-forward layer.
Prefix tuning \citep{Li2021} appends trainable prefix tokens to the input sequence, and prompt tuning \citep{Lester2021} perpends learnable continuous prompt vectors to the input embeddings.
LoRA \citep{Hu2021} decomposes the model weight matrix in the target layer into two trainable low-rank matrices.

\subsection{Quantization}
Another way to reduce computational requirements is by applying quantization techniques to LLMs to decrease the numerical precision of model parameters. 
The LLM.int8() \citep{Dettmers2022} quantizes model weights to 8-bit integers through a vector-wise quantization and mixed-precision decomposition. 
The QLoRA \citep{Dettmers2024} converts model weights to 4-bit integers by employing three techniques: 4-bit NormalFloat quantization, double quantization, and paged optimizers.

\begin{figure*}[h!]
    \centering
    \includegraphics[width=\textwidth]{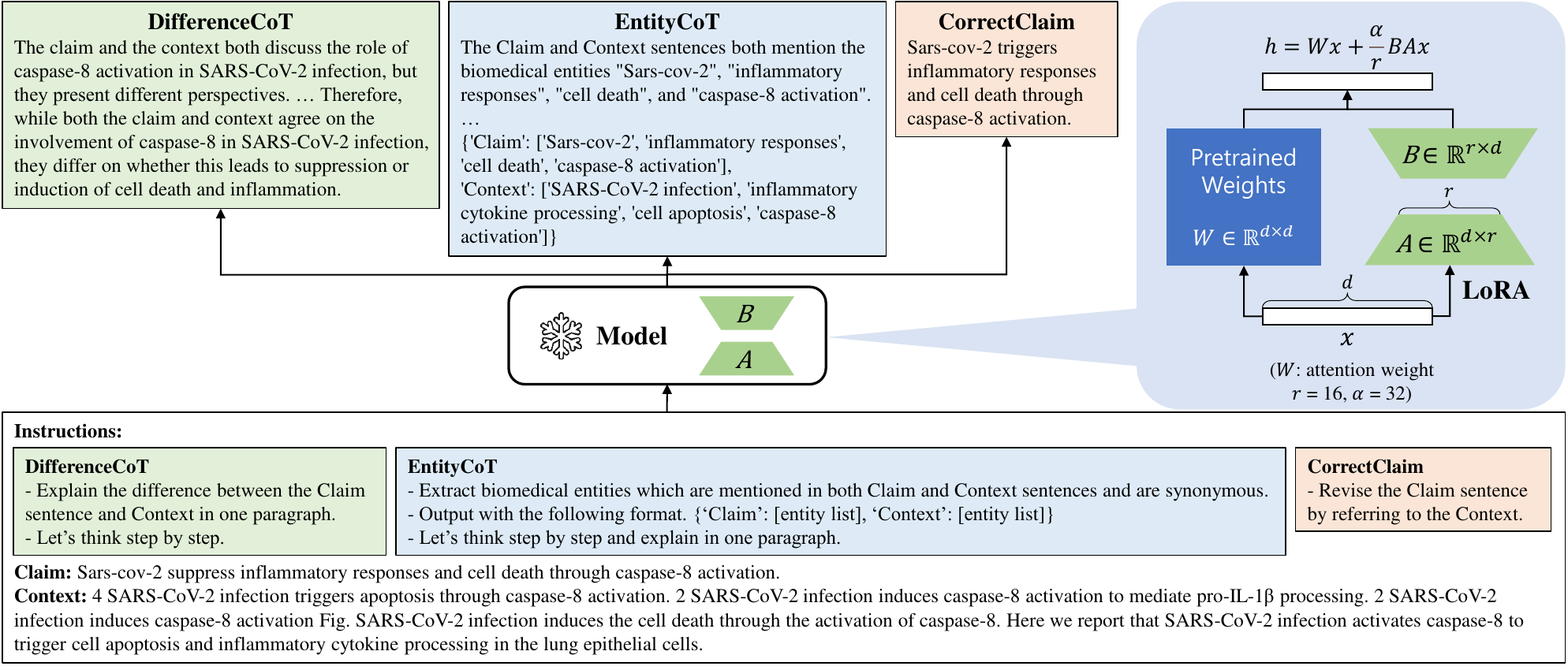}
    \caption{\label{reasoning-datasets}
Input and output examples of reasoning datasets for fine-tuning the generative model with LoRA. The LoRA consists of $A$ and $B$ weight matrices and exists in transformer layers.
}
\end{figure*}

\section{Methods}

\subsection{Reasoning Dataset Generation}

Determining the veracity of a claim requires identifying key entities and their relationships within the claim and evidence, and then analyzing the differences between them.
In this context, we hypothesize that identifying contrasting or common factors between the claim sentence and its corresponding evidence can assist the fact-checking model.
Therefore, we customize the three reasoning tasks for fact-checking: DifferenceCoT, EntityCoT, and CorrectClaim.
\begin{description}
\setlength{\itemsep}{0pt}
    \item[DifferenceCoT] is a task that explains the contextual differences between claim and evidence, including topic, level of detail, and relation.
    \item[EntityCoT] is a task that extracts synonymous biomedical entities that are concurrently present in both claim and evidence.
    \item[CorrectClaim] is a task that revises a given claim sentence based on the evidence.
\end{description}

Then, we construct the reasoning datasets using existing fact-checking datasets: COVID-Fact and SciFact.
The input of the reasoning datasets includes task instruction, claim, and evidence, as shown in Figure~\ref{reasoning-datasets}.
For DifferenceCoT and EntityCoT, we employ Chain-of-Thought (CoT) prompting \citep{Wei2022} and use GPT-4 API to generate the ground truth output.
For CorrectClaim, the ground truth output is the true claim for the given evidence.
The details and assessments of reasoning datasets are in Section 5.3 and Appendix A.

\subsection{Fine-tuning Reasoning LoRAs}

The next step is to fine-tune LoRAs for each task.
The lightweight LoRA module exists in transformer attention or feed-forward layers of the base model.
For each task $t\in\{1,2,3\}$, LoRA consists of a weight matrix $A_t\in \mathbb{R}^{d\times r}$ for down-projection of features to a smaller dimension $r$, and a weight matrix $B_t\in \mathbb{R}^{r\times d}$ for up-projection to the original dimension $d$, as depicted in Figure~\ref{reasoning-datasets}.
By freezing the base model weights and training only the LoRA weights, the fine-tuning process requires much fewer parameters.

Among encoder-decoder models, we use Flan-T5 due to its range of model size options and its strong performance in zero-shot, few-shot, and CoT \citep{Chung2022}.
In the Flan-T5 model, LoRA applies to the query and value projections of the encoder self-attention, decoder self-attention, and encoder-decoder attention layers.
For decoder-only models, we use LLaMA3 \citep{Dubey2024} as a small LLM and apply LoRA to the query, value, key, and output projections of the attention layers, up and down projections of the feed-forward layers, and the gate mechanism. 

\subsection{Connecting Reasoning LoRAs}
\begin{figure}[!b] 
    \centering
    \includegraphics[width=1\linewidth]{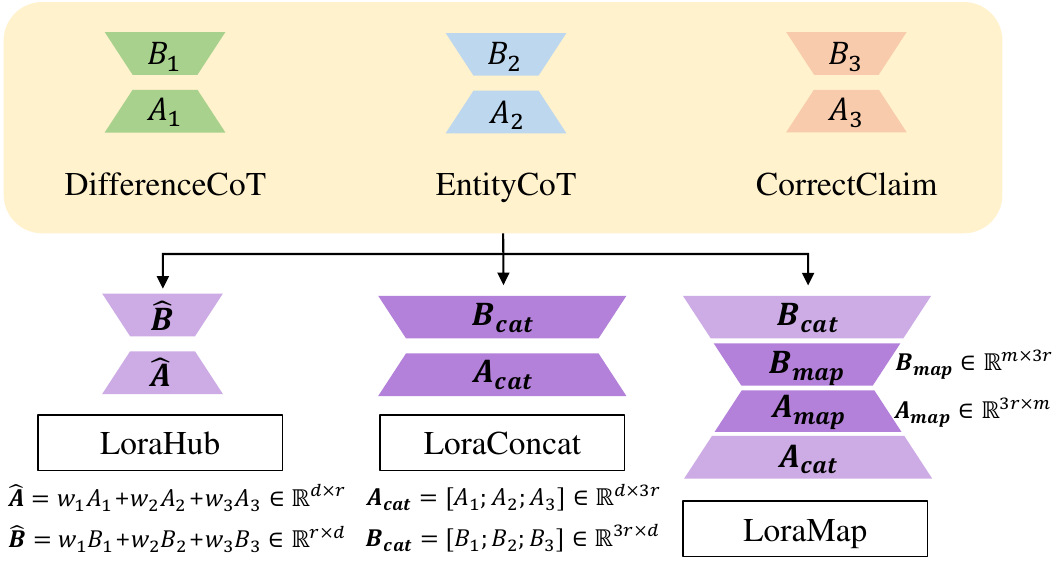}
    \caption{The comparison of LoraHub, LoraConcat, and LoraMap. Dark purple indicates trainable weights, and light purple represents fixed weights.}
    \label{loramap}
\end{figure}

The final step involves investigating methods to allocate and connect reasoning LoRAs, specifically LoraHub, LoraConcat, and LoraMap.
These methods differ in approaches for integrating the $A_t\in \mathbb{R}^{d\times r}$ and $B_t\in \mathbb{R}^{r\times d}$ matrices.
Figure~\ref{loramap} illustrates a visual comparison of these methods.

LoraHub combines LoRAs by computing the weighted sum of $A_t$ and $B_t$ matrices to generate $\hat{A}$ and $\hat{B}$ matrices.
\[
\hat{A} = \sum_{t=1}^n w_t A_t =  w_1 A_1 + \ldots + w_n A_n \in \mathbb{R}^{d \times r}
\]
\[
\hat{B} = \sum_{t=1}^n w_t B_t = w_1 B_1 + \ldots + w_n B_n \in \mathbb{R}^{r \times d}
\]
This method freezes all $A_{t}$ and $B_{t}$ matrices and learns only the coefficients $w_{t}$ for each LoRA.
The training is finding the optimal coefficients $w_{t}$ for a target dataset by a gradient-free approach. 
The original LoraHub randomly selects 20 LoRAs from approximately 200 LoRA modules trained on distinct tasks. 
Our LoraHub loads three reasoning LoRAs along with these 20 LoRA modules and learns 23 coefficients for each LoRA layer.

LoraConcat integrates LoRAs by concatenating the matrices $A_t$ and $B_t$ of the three reasoning LoRAs to produce $A_{cat}$ and $B_{cat}$ matrices.
\[
    A_{cat} = [A_{1};A_{2};A_{3}] \in \mathbb{R}^{d\times 3r}
\]
\[
    B_{cat} = [B_{1};B_{2};B_{3}] \in \mathbb{R}^{3r\times d}
\]
Then, we fine-tune the $A_{cat}$ and $B_{cat}$ matrices for the target dataset. As there are three LoRAs to combine, LoraConcat has $2 \times 3r \times d$ trainable parameters for each LoRA layer.

LoraMap not only concatenates the three reasoning LoRAs into $A_{cat}$ and $B_{cat}$ but also inserts the trainable matrices $A_{map}$ and $B_{map}$ between them. 
\[
    A_{map} \in \mathbb{R}^{3r\times m} \:\:,\:\: B_{map} \in \mathbb{R}^{m\times 3r}
\]

LoraMap freezes $A_{cat}$ and $B_{cat}$ to maintain specialized reasoning capabilities and learns the connection maps between them by fine-tuning only $A_{map}$ and $B_{map}$. 
Each LoRA layer has $2 \times 3r \times m$ trainable parameters.
We define the mapping dimension $m$ based on the ratio of trainable parameters to the total number of parameters in the model.
\[
    m = \frac{ratio \times num \: of \: total \: parameters}{3r \times num \: of \: trainable \: layers}
\]
The number of trainable layers refers to the total number of $A_{map}$ and $B_{map}$ layers.
The default value for $m$ is equivalent to $r$, allowing for adjustments to $m$ by modifying the ratio.

\begin{table}[h]
\centering
\resizebox{\columnwidth}{!}{%
\begin{tabular}{ccccc}
\hline
\textbf{Dataset} & \textbf{Partition} & \textbf{True Instances} & \textbf{False Instances} & \textbf{Total Instances} \\
\hline
\multirow{3}{*}{\textbf{COVID-Fact}} & Train & 1,018 & 1,018 & 2,036 \\ \cline{2-5}
                                     & Val   & 129   & 129   & 258   \\ \cline{2-5}
                                     & Test  & 128   & 128   & 256   \\ 
\hline
\multirow{3}{*}{\textbf{SciFact}}    & Train & 456   & 237   & 693   \\ \cline{2-5}
                                     & Val   & 124   & 64    & 188   \\ \cline{2-5}
                                     & Test  & 100   & 100   & 200   \\
\hline
\end{tabular}%
}
\caption{The statistics of the COVID-Fact and SciFact dataset for generating reasoning datasets}
\label{dataset_statistics}
\end{table}

\section{Experimental Results}

\subsection{Datasets}
We conduct experiments on COVID-Fact and SciFact datasets and extract data to construct reasoning datasets.
In the COVID-Fact dataset, each piece of evidence is associated with at least one true claim and one false claim.
For this dataset, we randomly select one true and one false claim per evidence, resulting in 2,550 instances.
In contrast, the SciFact dataset has three veracity labels: true, false, and not enough information. We include only the true and false claims and exclude claims without evidence, yielding 1,081 instances.
Additionally, since SciFact does not consistently offer both true and false claims for each evidence, we employ CoT prompting with the GPT-4 API to generate CorrectClaim. 
The statistics for these datasets are shown in Table~\ref{dataset_statistics}.

\subsection{Reasoning LoRAs}
Table~\ref{reasoning_lora_results} presents the results of three reasoning LoRAs, using BLEU \citep{Papineni2002}, ROUGE \citep{Lin2004,Lin2004_2}, and METEOR \citep{Banerjee2005} scores as lexical overlap-based metrics, and BERTscore \citep{Zhang2019} with the Longformer-base model as semantic embedding-based metrics.
Among models below 1B, we experiment with Flan-T5-large (787M) and for models above 1B, we use Flan-T5-xxl (11B) quantizied with LLM.int8() and LLaMA3 (8B) quantizied with QLoRA.

In the zero-shot setting, the base model performs reasoning tasks without fine-tuning, resulting in poor scores. 
Fine-tuning LoRA on each reasoning dataset improves performance on all metrics.
LLaMA3 outperforms other models on DifferenceCoT and EntityCoT of COVID-Fact datasets, whereas it has weaknesses on the CorrectClaim.
We suppose this is primarily due to its lower zero-shot performance on CorrectClaim, which also impacts its fine-tuning performance.
All experiments use a fixed seed 42 for reproducibility, and the experimental details are in Appendix B.1.

\begin{table*}[h]
\centering
\resizebox{\textwidth}{!}{%
\begin{tabular}{ccccccccccc}
\hline
\textbf{Dataset} & \textbf{\begin{tabular}[c]{@{}c@{}}Reasoning \\ LoRA\end{tabular}} & \textbf{\begin{tabular}[c]{@{}c@{}}Base model \\ (total params)\end{tabular}} & \textbf{\begin{tabular}[c]{@{}c@{}}Setting \\ (trainable params)\end{tabular}} & \textbf{BLEU} & \textbf{ROUGE-1} & \textbf{ROUGE-2} & \textbf{ROUGE-L} & \textbf{ROUGE-Lsum} & \textbf{METEOR} & \textbf{BERTscore} \\ \hline

\multirow{18}{*}{COVID-Fact} &

\multirow{6}{*}{DifferenceCoT} & 

\multirow{2}{*}{\begin{tabular}[c]{@{}c@{}}Flan-T5-large \\ (787M)\end{tabular}} & Zero-shot & 0.0023 & 0.2173 & 0.1326 & 0.1815 & 0.2011 & 0.1047 & 0.8563 \\ \cline{4-11} 
 &  &  & LoRA finetuning (4M) & 0.3588 & 0.6676 & 0.4206 & 0.5045 & 0.6310 & 0.5255 & 0.9275 \\ \cline{3-11} 
 
 & & \multirow{2}{*}{\begin{tabular}[c]{@{}c@{}}Flan-T5-xxl \\ (11B)\end{tabular}}& Zero-shot & 0.0012 & 0.2034 & 0.1298 & 0.1718 & 0.1875 & 0.0945 & 0.8545 \\ \cline{4-11} 
 &  & & qLoRA finetuning (18M) & 0.3764 & 0.6822 & 0.4446 & 0.5192 & 0.6444 & 0.5245 & 0.9315 \\ \cline{3-11} 
 
& & \multirow{2}{*}{\begin{tabular}[c]{@{}c@{}}LLaMA3\\ (8B)\end{tabular}} & Zero-shot & 0.2057 & 0.6097 & 0.3180 & 0.3969 & 0.5655 & 0.4424 & 0.9092 \\ \cline{4-11} 
 &  & & qLoRA finetuning (41M) & \textbf{0.3889} & \textbf{0.6980} & \textbf{0.4620} & \textbf{0.5319} & \textbf{0.6597} & \textbf{0.5582} & \textbf{0.9337} \\ \cline{2-11}
 
& \multirow{6}{*}{EntityCoT} & 
 
 \multirow{2}{*}{\begin{tabular}[c]{@{}c@{}}Flan-T5-large \\ (787M)\end{tabular}} & Zero-shot & 0 & 0.0539 & 0.0201 & 0.0533 & 0.0526 & 0.0289 & 0.7997 \\ \cline{4-11} 
 & & & LoRA finetuning (4M) & 0.3885 & 0.6755 & 0.4533 & 0.5548 & 0.6397 & 0.5969 & 0.9240 \\ \cline{3-11} 
 
& & \multirow{2}{*}{\begin{tabular}[c]{@{}c@{}}Flan-T5-xxl \\ (11B)\end{tabular}} & Zero-shot & 0 & 0.0444 & 0.0238 & 0.0444 & 0.0442 & 0.0097 & 0.7903 \\ \cline{4-11} 
 & & & qLoRA finetuning (18M) & 0.3805 & 0.6680 & 0.4505 & 0.5500 & 0.6356 & 0.5881 & 0.9223 \\ \cline{3-11} 
 
& & \multirow{2}{*}{\begin{tabular}[c]{@{}c@{}}LLaMA3\\ (8B)\end{tabular}} & Zero-shot & 0.1799 & 0.5200 & 0.2451 & 0.3221 & 0.4826 & 0.4094 & 0.8850 \\ \cline{4-11} 
 & & & qLoRA finetuning (41M) & \textbf{0.4115} & \textbf{0.6819} & \textbf{0.4604} & \textbf{0.5577} & \textbf{0.6513} & \textbf{0.6092} & \textbf{0.9273} \\ \cline{2-11}

& \multirow{6}{*}{CorrectClaim} & 

\multirow{2}{*}{\begin{tabular}[c]{@{}c@{}}Flan-T5-large \\ (787M)\end{tabular}} & Zero-shot & 0.3636 & 0.6839 & 0.5714 & 0.6618 & 0.6636 & 0.6591 & 0.9349 \\ \cline{4-11} 
 & & & LoRA finetuning (4M) & \textbf{0.9257} & \textbf{0.9722} & \textbf{0.9437} & \textbf{0.9721} & \textbf{0.9721} & \textbf{0.9682} & \textbf{0.9944} \\ \cline{3-11} 
 
& & \multirow{2}{*}{\begin{tabular}[c]{@{}c@{}}Flan-T5-xxl \\ (11B)\end{tabular}} & Zero-shot & 0.5212 & 0.8102 & 0.7251 & 0.7985 & 0.7983 & 0.7821 & 0.9565 \\ \cline{4-11} 
 & & & qLoRA finetuning (18M) & 0.9227 & 0.9700 & 0.9389 & 0.9695 & 0.9696 & 0.9662 & 0.9943 \\ \cline{3-11} 
 
& & \multirow{2}{*}{\begin{tabular}[c]{@{}c@{}}LLaMA3\\ (8B)\end{tabular}} & Zero-shot & 0.054 & 0.3248 & 0.2019 & 0.2882 & 0.2907 & 0.4766 & 0.8695 \\ \cline{4-11} 
 & & & qLoRA finetuning (41M) & 0.3664 & 0.9676 & 0.9377 & 0.9671 & 0.9670 & 0.9562 & 0.9890 \\ \hline

\multirow{6}{*}{SciFact} &

\multirow{2}{*}{DifferenceCoT} 

& \multirow{2}{*}{\begin{tabular}[c]{@{}c@{}}LLaMA3\\ (8B)\end{tabular}} & Zero-shot & 0.2185 & 0.5873 & 0.3135 & 0.3966 & 0.5490 & 0.4362 & 0.9096 \\ \cline{4-11} 
 & & & qLoRA finetuning (41M) & \textbf{0.3663} & \textbf{0.6799} & \textbf{0.4521} & \textbf{0.5261} & \textbf{0.6513} & \textbf{0.5610} & \textbf{0.9357} \\ \cline{2-11}

& \multirow{2}{*}{EntityCoT} 

& \multirow{2}{*}{\begin{tabular}[c]{@{}c@{}}LLaMA3\\ (8B)\end{tabular}} & Zero-shot & 0.1839 & 0.5060 & 0.2277 & 0.3271 & 0.4714 & 0.4547 & 0.8867 \\ \cline{4-11} 
 & & & qLoRA finetuning (41M) & \textbf{0.4200} & \textbf{0.6766} & \textbf{0.4503} & \textbf{0.5397} & \textbf{0.6412} & \textbf{0.6246} & \textbf{0.9272} \\ \cline{2-11}

& \multirow{2}{*}{CorrectClaim} 

& \multirow{2}{*}{\begin{tabular}[c]{@{}c@{}}LLaMA3\\ (8B)\end{tabular}} & Zero-shot & 0.2627 & 0.5984 & 0.3701 & 0.4358 & 0.5582 & 0.4356 & 0.9210 \\ \cline{4-11} 
 & & & qLoRA finetuning (41M) & \textbf{0.4055} & \textbf{0.6861} & \textbf{0.4970} & \textbf{0.5464} & \textbf{0.6571} & \textbf{0.6308} & \textbf{0.9412} \\ \hline
\end{tabular}%
}
\caption{The evaluation results of three reasoning LoRAs on the COVID-Fact and SciFact test dataset.
The bold text indicates the best performance for each reasoning LoRA.}
\label{reasoning_lora_results}
\end{table*}

\begin{table*}[h!t]
\centering
\resizebox{\textwidth}{!}{%
\begin{tabular}{ccccccc}
\hline
\textbf{Model} & \textbf{Reasoning LoRA} & \textbf{Fact-checking setting} & \textbf{\# Training instances} & \textbf{Macro-precision} & \textbf{Macro-recall} & \textbf{Macro-f1} \\ \hline
\multirow{10}{*}{\begin{tabular}[c]{@{}c@{}}Flan-T5-\\ large\end{tabular}} & --- & Zero-shot & 0 & 0.7819 & 0.6133 & 0.5453 \\ \cline{2-7} 
 & \multirow{3}{*}{\begin{tabular}[c]{@{}c@{}}base20 + DifferenceCoT + \\ EntityCoT + ClaimCorrection\end{tabular}} & \multirow{3}{*}{LoraHub} & 50 & \textbf{0.6664} & \textbf{0.6344} & 0.6133 \\
 &  &  & 200 & 0.6643 & 0.6340 & \textbf{0.6145} \\
 &  &  & 2,036* & 0.6589 & 0.6254 & 0.6030 \\ \cline{2-7} 
 & \multirow{3}{*}{\begin{tabular}[c]{@{}c@{}}DifferenceCoT + \\ EntityCoT + \\ ClaimCorrection\end{tabular}} & \multirow{3}{*}{LoraConcat (14M)} & 100 & 0.8087 & 0.7828 & 0.7782 \\
 &  &  & 1,000 & \textbf{0.8334} & \textbf{0.8152} & \textbf{0.8126} \\
 &  &  & 2,036* & 0.8184 & 0.8082 & 0.7910 \\ \cline{2-7} 
 & \multirow{3}{*}{\begin{tabular}[c]{@{}c@{}}DifferenceCoT + \\ EntityCoT + \\ ClaimCorrection\end{tabular}} & \multirow{3}{*}{LoraMap (0.22M)} & 100 & 0.7527 & 0.6961 & 0.6755 \\
 &  &  & 1,000 & 0.8052 & 0.8015 & 0.8010 \\
 &  &  & 2,036* & \textbf{0.8302} & \textbf{0.8246} & \textbf{0.8239}  \\ \hline
\end{tabular}%
}
\caption{The evaluation results of the Flan-T5-large model on the COVID-Fact test dataset.
In the fact-checking settings, the values in parenthesis indicates the number of trainable parameters.
Bold text highlights the best performance and * denotes the size of the entire training data.
}
\label{table3}
\end{table*}

\subsection{Integrating LoRAs for Fact-checking}

Given the prompt ``{\em What is the class of the Claim by referring to the Context? Choose only from TRUE or FALSE.}'' along with the claim and context, we align the model output to follow the zero-shot output format. ``{\em The claim is TRUE/FALSE}'' for Flan-T5 models and ``{\em Based on the context, the given claim is TRUE/FALSE}'' for the LLaMA3 model.

\subsubsection{Results of Small Language Model}
The performance of the Flan-T5-large on the COVID-Fact test dataset is shown in Table~\ref{table3}.
In the zero-shot setting, the model predominantly predicts TRUE with an f1 score of 0.5453. 
The key result is a comparison of LoraHub, LoraConcat, and LoraMap. 
We experiment with various training instances, and Table~\ref{table3} presents the best result among 10-shot, 20-shot, 50-shot, and 100-shot, the best result among 200-shot, 500-shot, and 1000-shot, and the result when using the entire dataset.
Training with more than 100 instances is to identify the minimum number of instances required to achieve satisfactory performance in LoraConcat and LoraMap.
To provide statistically reliable results, all metric scores are the average of ten repeated experiments, each with a fixed seed (42, 64, 128, 256, 512, 1024, 2048, 4096, 8192, 16384).

LoraHub achieves the highest f1-score at 200-shot, and its performance does not increase as the number of training data increases. Although training LoraHub with less than 100 examples is feasible, its performance is suboptimal.
In contrast, LoraConcat and LoraMap generally demonstrate improved f1-scores as training instances increase. LoraConcat yields the best f1-score of 0.8126 at 1000-shot, and LoraMap with default $m$(16) achieves the best f1-score of 0.8239 when using all instances. LoraMap achieves statistically significant superior performance even with fewer trainable parameters than LoraConcat, with a p-value of 0.03756.

\begin{table*}[h]
\resizebox{\textwidth}{!}{%
\begin{tabular}{ccccccc}
\hline
\textbf{Model} & \textbf{Reasoning LoRA} & \textbf{\# Training instances} & \textbf{Macro-precision} & \textbf{Macro-recall} & \textbf{Macro-f1} \\ \hline
 &  & 0 & 0.7643 & 0.6094 & 0.5423 \\ \cline{3-6} 
 & \multirow{-2}{*}{base20} & 20 & 0.7529 & 0.6836 & \textbf{0.6603} \\ \cline{2-6} 
 &  & 0 & 0.6900 & 0.6758 & \textbf{0.6696} \\ \cline{3-6} 
 & \multirow{-2}{*}{DifferenceCoT + EntityCoT + ClaimCorrection} & 10 & 0.6889 & 0.6133 & 0.5703 \\ \cline{2-6} 
 &  & 0 & 0.7647 & 0.6367 & \textbf{0.5868} \\ \cline{3-6} 
 & \multirow{-2}{*}{base3} & 10 & 0.7771 & 0.5977 & 0.5199 \\ \cline{2-6} 
 &  & 0 & 0.7807 & 0.6094 & 0.5390 \\ \cline{3-6} 
\multirow{-8}{*}{Flan-T5-large} & \multirow{-2}{*}{base20 + DifferenceCoT + EntityCoT + ClaimCorrection} & 50 & 0.6833 & 0.6797 & \textbf{0.6781} \\ \hline
\end{tabular}%
}
\caption{LoraHub results for the COVID-Fact test dataset depending on the selection of LoRAs.
}
\label{table4}
\end{table*}

\begin{table*}[h]
\resizebox{\textwidth}{!}{%
\begin{tabular}{ccccccc}
\hline
\textbf{Model} & \textbf{Reasoning LoRA} & \textbf{\# Trainable parameter} & \textbf{Macro-precision} & \textbf{Macro-recall} & \textbf{Macro-f1} \\ \hline
\multirow{4}{*}{Flan-T5-large} & \begin{tabular}[c]{@{}c@{}c}DifferenceCoT + EntityCoT\end{tabular} & 147,456 & 0.7965 & 0.7852 & 0.7831 \\ \cline{2-6} 
 & \begin{tabular}[c]{@{}c@{}}DifferenceCoT + ClaimCorrection\end{tabular} & 147,456 & 0.7969 & 0.7969 & 0.7969 \\ \cline{2-6} 
 & \begin{tabular}[c]{@{}c@{}}EntityCoT + ClaimCorrection\end{tabular} & 147,456 & 0.7723 & 0.7656 & 0.7642 \\ \cline{2-6} 
 & \begin{tabular}[c]{@{}c@{}}DifferenceCoT + EntityCoT + ClaimCorrection\end{tabular} & 221,184 & \textbf{0.8347} & \textbf{0.8281} & \textbf{0.8273} \\ \hline
\end{tabular}%
}
\caption{LoraMap results for the COVID-Fact test dataset depending on the selection of reasoning LoRAs.
}
\label{table5}
\end{table*}

In terms of parameter efficiency, LoraHub has 3,312 (23$\times$144) trainable coefficients out of a total of 787M parameters, as the model consists of 144 layers, each with 23 LoRAs. There are 14M (1024$\times$48$\times$288) trainable parameters out of 797.3M parameters when using LoraConcat and 0.22M (48$\times$16$\times$288) trainable parameters out of 797.5M parameters when using LoraMap. We conduct all experiments with two RTX 3090 GPUs and compare the training and inference time. Training on all COVID-Fact train datasets takes 1 hour 44 minutes for LoraHub, 5 hours 7 minutes for LoraConcat, and 4 hours 14 minutes for LoraMap. For each test instance, inferencing takes less than 0.3 seconds for LoraHub and less than 0.5 seconds for LoraConcat and LoraMap.

\subsubsection{Ablation Study}
We further compare the results depending on the selection of LoRAs.
Table~\ref{table4} exhibits the results of LoraHub across different training instances, presenting zero-shot and the best result among 10-shot, 20-shot, 50-shot, and 100-shot learning.
All experiments use a fixed seed of 42. 
The original LoraHub uses 20 random LoRAs (base20) and shows improvement after fine-tuning 20 coefficients per layer.
When employing only three reasoning LoRAs, the zero-shot performance is higher than that of base20. However, the performance does not improve while fine-tuning due to the difficulty of training only with three coefficient weights. We also experiment with three random LoRAs (base3) to verify this, and the results demonstrate the same tendency to struggle with fine-tuning. Consequently, we keep base20 and add three reasoning LoRAs, yielding the best macro f1 score. 

Additionally, LoraHub outputs coefficients that indicates the impact of each LoRA module after training. The coefficients for the three reasoning LoRAs are all close to 0.5. Similarly, four out of the 20 base modules, mostly trained for question-answering, also show 0.5 and the remaining modules show values close to zero or negative. The coefficients confirm that our reasoning LoRAs play an important role in fact-checking. 

Table~\ref{table5} shows the results of an ablation study on LoraMap to demonstrate the effectiveness of each LoRA. 
All experiments use the entire training dataset with a fixed seed 42.
Removing each LoRA degrades the macro-f1 score, and the most influential one is DifferenceCoT LoRA, which exhibits the largest performance decrease. 
DifferenceCoT, ClaimCorrection, and EntityCoT are most influential in that order, indicating that the tasks identifying and correcting differences between claims and context are more beneficial than finding synonymous entities within it.

\begin{table*}[h!t]
\centering
\resizebox{\textwidth}{!}{%
\begin{tabular}{ccccccccc}
\hline
\textbf{Dataset} & \textbf{Model} & \textbf{Reasoning LoRA} & \textbf{Fact-checking setting} & \textbf{ratio$\times$100} & \textbf{$m$} & \textbf{Macro-precision} & \textbf{Macro-recall} & \textbf{Macro-f1} \\ \hline

\multirow{17}{*}{COVID-Fact} &

GPT-4 & --- & Zero-shot & --- & --- & 0.7426 & 0.7070 & 0.6959 \\ \cline{2-9}

& \multirow{9}{*}{\begin{tabular}[c]{@{}c@{}}Flan-T5-\\ xxl (11B)\end{tabular}} & --- & Zero-shot & --- & --- & 0.7392 & 0.7109 & 0.7021 \\ \cline{3-9}

&  & {\begin{tabular}[c]{@{}c@{}}DifferenceCoT + \\ EntityCoT + ClaimCorrection\end{tabular}} & {LoraConcat (56M)} & --- & --- & 0.8907 & 0.8906 & 0.8906 \\ \cline{3-9}

&   & \multirow{6}{*}{\begin{tabular}[c]{@{}c@{}}DifferenceCoT + \\ EntityCoT + \\ ClaimCorrection\end{tabular}} & {LoraMap (5.5M)} & 0.05 & 400 & 0.8879 & 0.8867 & 0.8866 \\ \cline{4-9}
&   &  & {LoraMap (4.4M)} & 0.04 & 320 & \textbf{0.8947} & \textbf{0.8945} & \textbf{0.8945} \\ \cline{4-9}
&   & & {LoraMap (3.3M)} & 0.03 & 240 & 0.8837 & 0.8828 & 0.8827 \\ \cline{4-9}
&   & & {LoraMap (2.2M)} & 0.02 & 160 & 0.8671 & 0.8633 & 0.8629 \\ \cline{4-9}
&   & & {LoraMap (1.1M)} & 0.01 & 80 & 0.8565 & 0.8555 & 0.8554 \\ \cline{4-9}
&   & & {LoraMap (0.22M)} & 0.002 & 16 & 0.8043 & 0.7969 & 0.7956 \\ \cline{4-9}
   \cline{2-9}
   
& \multirow{9}{*}{\begin{tabular}[c]{@{}c@{}}LLaMA3\\ (8B)\end{tabular}} & --- & Zero-shot & --- & --- & 0.6988 & 0.6172 & 0.5734 \\ \cline{3-9}

&  & {\begin{tabular}[c]{@{}c@{}}DifferenceCoT + \\ EntityCoT + ClaimCorrection\end{tabular}} & {LoraConcat (125M)} & --- & --- & 0.8076 & 0.8008 & 0.7997 \\ \cline{3-9}

&   & \multirow{6}{*}{\begin{tabular}[c]{@{}c@{}}DifferenceCoT + \\ EntityCoT + \\ ClaimCorrection\end{tabular}} & {LoraMap (4.1M)} & 0.05 & 192 & \textbf{0.8210} & \textbf{0.8203} & \textbf{0.8202} \\ \cline{4-9}
&   &  & {LoraMap (3.0M)} & 0.04 & 144 & 0.7824 & 0.7812 & 0.7810 \\ \cline{4-9}
&   & & {LoraMap (2.4M)} & 0.03 & 112 & 0.7802 & 0.7773 & 0.7768 \\ \cline{4-9}
&   & & {LoraMap (1.7M)} & 0.02 & 80 & 0.7270 & 0.7171 & 0.7139 \\ \cline{4-9}
&   & & {LoraMap (0.6M)} & 0.01 & 32 & 0.7123 & 0.7016 & 0.6977 \\ \cline{4-9}
&   & & {LoraMap (0.3M)} & 0.005 & 16 & 0.7093 & 0.7031 & 0.7009 \\ \cline{4-9}
   \hline

\multirow{9}{*}{SciFact} &

\multirow{9}{*}{\begin{tabular}[c]{@{}c@{}}LLaMA3\\ (8B)\end{tabular}} & --- & Zero-shot & --- & --- & 0.8111 & 0.8100 & 0.8098 \\ \cline{3-9}

& & {\begin{tabular}[c]{@{}c@{}}DifferenceCoT + \\ EntityCoT + ClaimCorrection\end{tabular}} & {LoraConcat (125M)} & --- & --- & 0.8347 & 0.8250 & 0.8237 \\ \cline{3-9}
  
& & \multirow{6}{*}{\begin{tabular}[c]{@{}c@{}}DifferenceCoT + \\ EntityCoT + \\ ClaimCorrection\end{tabular}} & {LoraMap (4.1M)} & 0.05 & 192 & \textbf{0.8802} & \textbf{0.8800} & \textbf{0.8800} \\ \cline{4-9}
&   &  & {LoraMap (3.0M)} & 0.04 & 144 & 0.8659 & 0.8650 & 0.8649 \\ \cline{4-9}
&   & & {LoraMap (2.4M)} & 0.03 & 112 & 0.8653 & 0.8650 & 0.8650 \\ \cline{4-9}
&   & & {LoraMap (1.7M)} & 0.02 & 80 & 0.8659 & 0.8650 & 0.8649 \\ \cline{4-9}
&   & & {LoraMap (0.6M)} & 0.01 & 32 & 0.8606 & 0.8600 & 0.8599 \\ \cline{4-9}
&   & & {LoraMap (0.3M)} & 0.005 & 16 & 0.8606 & 0.8600 & 0.8599 \\ \cline{4-9}
   \hline

\end{tabular}%
}
\caption{The results of LLMs on the COVID-Fact and SciFact test datasets.
In the fact-checking settings, the values in parenthesis indicate the number of trainable parameters.
The bold text represents the best result for each model.
}
\label{table6}
\end{table*}

\subsubsection{Applicability to LLMs}
Table~\ref{table6} shows the performance of LLMs on the COVID-Fact and SciFact test datasets.
We compare the performance of zero-shot CoT with the fine-tuned LoraConcat and LoraMap using all training instances.
In the COVID-Fact dataset, Flan-T5-xxl with LoraMap (4.4M) achieves the best overall scores, potentially due to its substantial model size.
For the Flan-T5-xxl model, LoraMap (4.4M) outperforms LoraConcat (56M), and for the LLaMA3 model, LoraMap (4.1M) surpasses LoraConcat (125M).
Likewise, in the SciFact dataset, LLaMA3 with LoraMap (4.1M) exceeds the performance of LoraConcat (125M).

The results demonstrate that LoraMap consistently outperforms LoraConcat even with significantly fewer trainable parameters.
In the case of using LoraMap with LLMs, the default value of $m$ at 16 does not provide sufficient performance.
Therefore, by adjusting the ratio of trainable to total parameters, it is feasible to increase $m$ to a higher dimension.
Across all models, the macro-f1 generally improves as the size of LoraMap increases.

Table~\ref{efficiency} compares trainable parameters, total parameters, training time, and inference time for each test instance. 
The models using LoraMap have higher total parameters than those using LoraConcat, but have fewer trainable parameters.
For the Flan-T5-xxl model, LoraMap also reduces training and inference times compared to LoraConcat.
In contrast, while LoraMap considerably shortens the inference time for the LLaMA3 model, it requires a longer training time because LoraMap tends to train for more epochs than LoraConcat.
The details of experimental settings are in Appendix B.2.

\begin{table}[b]
\centering
\resizebox{\columnwidth}{!}{%
\begin{tabular}{ccccc}
\hline
\textbf{Dataset} & \textbf{Model} & \textbf{Fact-checking Setting} & \textbf{Total params} & \textbf{Train/Test Time} \\
\hline
\multirow{4}{*}{\begin{tabular}[c]{@{}c@{}}COVID- \\ Fact\end{tabular}} & 
\multirow{2}{*}{\begin{tabular}[c]{@{}c@{}}Flan-T5-xxl \\ (11B)\end{tabular}}
& LoraConcat (56M) & 11.191B & 17h 5m / 3s \\ \cline{3-5}
& & LoraMap (4.4M) & 11.196B & 15h 22m / 2s \\ \cline{2-5}
& \multirow{2}{*}{\begin{tabular}[c]{@{}c@{}}LLaMA3 \\ (8B)\end{tabular}} & LoraConcat (125M) & 8.156B & 2h 43m / 22s \\ \cline{3-5}
& & LoraMap (4.1M) & 8.160B & 5h 6m / 5s \\ 
\hline
\multirow{2}{*}{SciFact} & 
\multirow{2}{*}{\begin{tabular}[c]{@{}c@{}}LLaMA3 \\ (8B)\end{tabular}} 
& LoraConcat (125M) & 8.156B & 1h 41m / 41s \\ \cline{3-5}
& & LoraMap (4.1M) & 8.160B & 1h 50m / 5s \\ \cline{2-5}
\hline
\end{tabular}%
}
\caption{Efficiency in terms of parameter and time.}
\label{efficiency}
\end{table}

\section{Discussion}
\subsection{Design Motivation of LoraMap}
The experimental findings highlight the significance of the integration strategies of multiple reasoning LoRAs. The main motivation for the LoraMap architecture is that the LoraHub linearly adds all trained LoRA weights. This linear approach can diminish the impact of individual matrix values due to the averaging effect, especially when there is a significant variation in weights according to the distinct roles of LoRAs.
We believe that in the human brain, training does not occur through linear addition but rather domain-specific training to enhance the brain’s functions for a specific task. 

Additionally, the LoraConcat architecture may experience a loss of reasoning capability due to the catastrophic forgetting problem as the concatenated LoRA matrices undergo further fine-tuning.
To address this, we design LoraMap to preserve these matrices and learn only the connections between LoRAs to support decision-making from diverse reasoning perspectives. As each brain region possesses different knowledge and functionalities, establishing interconnections among them would be important. Therefore, we train $A_{map}$and $B_{map}$ matrices while maintaining areas for each distinct function, which are $A_{cat}$ and $B_{cat}$ matrices.

According to the design, the differences in the number of trainable parameters can influence performance. 
When combining three LoRAs, LoraHub only learns three coefficients per LoRA layer, which may not be sufficient for complex tasks.
In contrast, LoraConcat learns ${2\times d\times 3r}$ parameters, and LoraMap learns ${2\times 3r\times m}$ parameters for each LoRA layer, resulting in better performance than LoraHub.
Additionally, increasing the number of LoRAs can lead to substantial parameter growth with a fixed $m$.
However, we can adjust $m$ of LoraMap to modify the number of trainable parameters for efficient training.

\subsection{Case Study}
The cases where the output of LoraMap is correct, but zero-shot and LoraConcat are incorrect on the COVID-Fact and SciFact datasets are in Figure\ref{app_case_study_covidfact_llama} and \ref{app_case_study_scifact_llama}, respectively.
Even though the fact-checking labels only classify claims, LoraMap generates explanations along with classifications, possibly because it uses reasoning LoRAs.
The details are in Appendix C.

\subsection{Applying LoraMap to Other Tasks}

LoraMap can be applied to other tasks but it requires some modifications. First, the reasoning LoRAs to integrate should be relevant to the downstream task. For a question-answering task, where a question and context are given for answering, the DifferenceCoT and EntityCoT could work as helper tasks, whereas ClaimCorrect may not be appropriate. 

Second, the inclusion and training of new reasoning LoRA is available. 
For any new helper task, it is necessary to train a new LoRA and then subsequently train the LoraHub, LoraConcat, and LoraMap models.
All these models need retraining to adjust the coefficient or corresponding matrix weights. 
We also considered predicting the triplets (entity-relation-entity) from the claim and context, but the poor performance of GPT-4 led to its exclusion from this study.

Third, if the researcher customizes five reasoning LoRAs, the LoraMap dimension changes. The original LoraMap matrices consist of $A_{cat}$ ($\mathbb{R}^{d\times 3r}$), $B_{cat}$ ($\mathbb{R}^{3r\times d}$), $A_{map}$ ($\mathbb{R}^{3r\times m}$), $B_{map}$ ($\mathbb{R}^{m\times 3r}$), and when employing five LoRAs, the dimension of $3r$ in each matrix changes to $5r$. 
As the number of LoRAs increases, so does the computational cost. 
Therefore, selecting LoRAs relevant to the downstream task and adjusting $m$ would be necessary.

\subsection{Reasoning Dataset Assessment}

Two graduate students specializing in biomedical engineering perform a manual assessment to evaluate the quality of reasoning datasets generated by GPT-4.
We randomly select 100 instances from each dataset for evaluation.
The DifferenceCoT dataset shows an accuracy of 0.93, and the EntityCoT dataset shows an accuracy of 0.89. 
Additionally, we compute Cohen’s kappa to assess the inter-rater reliability of manual evaluations. 
The kappa values for DifferenceCoT and EntityCoT are 0.8465 and 0.8629, respectively. 
Then, an emergency medicine clinician with over five years of experience reviews and confirms the dataset.

In DifferenceCoT, GPT-4 struggles with differentiating between claims and context, especially when dealing with numerical values.
For example, it overlooks differences such as the claim mentioning two hours while the context refers to two weeks or the claim mentioning 1,000 people while the context refers to at least 1 percent of the population.
Additionally, GPT-4 faces challenges due to a lack of biomedical knowledge, confusing bacterial viromes with human viromes.

In EntityCoT, GPT-4 incorrectly identifies distinct biomedical entities as synonymous, such as equating the `n gene of sars-cov-2' with the `n gene assay' and confusing `covid-19 infection' from the claim with `covid-19 vaccine prospects' from the context.
It also fails to recognize some synonymous entities, such as `sars-cov-2' in the claim and `COVID-19' in the context. 
In one case, GPT-4 hallucinates by identifying an entity in the claim that is not present in the context.

\section{Conclusion}
This paper investigates methods for integrating multiple reasoning LoRAs. 
We generate three reasoning datasets and fine-tune individual LoRAs to facilitate inference from different perspectives. 
Subsequently, we introduce LoraMap, which learns the connection map between reasoning LoRAs.
The fact-checking results show that LoraMap demonstrates superior performance and efficiency on the Flan-T5 and LLaMA3 models.
Future work could explore generating not only claim classifications but also explanations, applying LoraMap to other tasks, and developing a method to select relevant LoRAs for a given task.
We anticipate that this paper will pave the way for approaches to establish connections between LoRAs.

\section{Limitations}

Researchers should use GPT-4 or other APIs to generate reasoning datasets, which may incur some costs. Additionally, it is necessary to develop methods for evaluating the quality of GPT-4 reasoning and filtering data for training.

This paper focuses on fact-checking a single claim. In real-world scenarios, the outputs from LLM need to be verified. As traditional fact-checking models mostly verify a single claim, some research transforms LLM output into multiple claims. By verifying each claim and averaging their veracity, we can assess the reliability of the LLM outputs.

Our model is unsuitable for cases where only claims are present without evidence. In such cases, it is necessary to search for and provide appropriate evidence. There was no need to search for evidence in this work, as we use the claims with corresponding evidence. Furthermore, our model cannot make integrated judgments about multiple pieces of evidence.

Finally, it would be beneficial to examine LoraConcat and LoraMap across various open-source LLMs and other large fact-checking datasets within the biomedical and health domains.

\label{sec:bibtex}

\bibliography{custom}

\appendix

\section{Details of Datasets}

\subsection{Examples of Reasoning Datasets}

\begin{figure*}[]
    \centering
    \includegraphics[width=\textwidth]{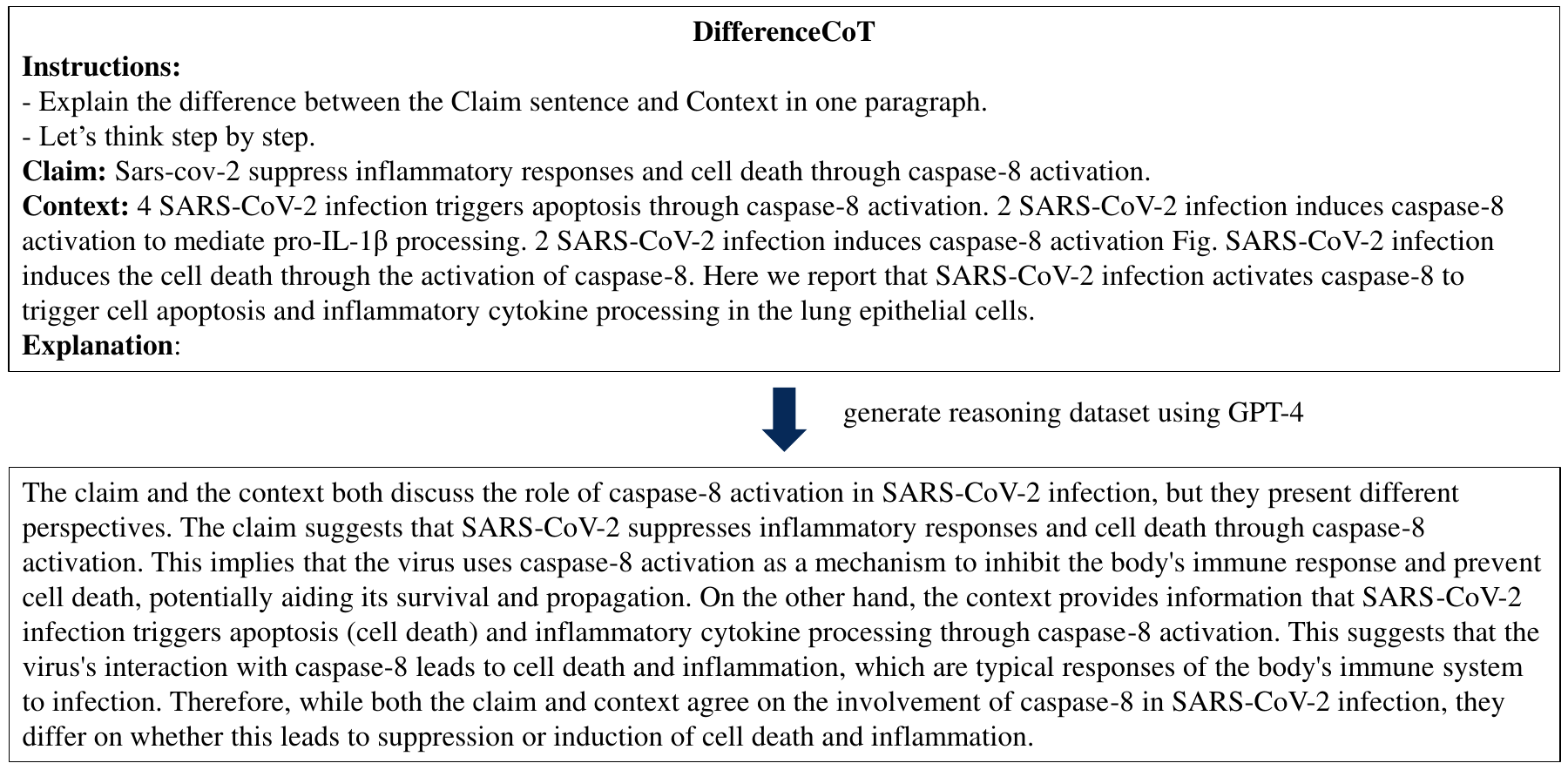}
    \caption{An example of DifferenceCoT dataset on the COVID-Fact.}
    \label{app_DifferenceCoT}
\end{figure*}

\begin{figure*}[]
    \centering
    \includegraphics[width=\textwidth]{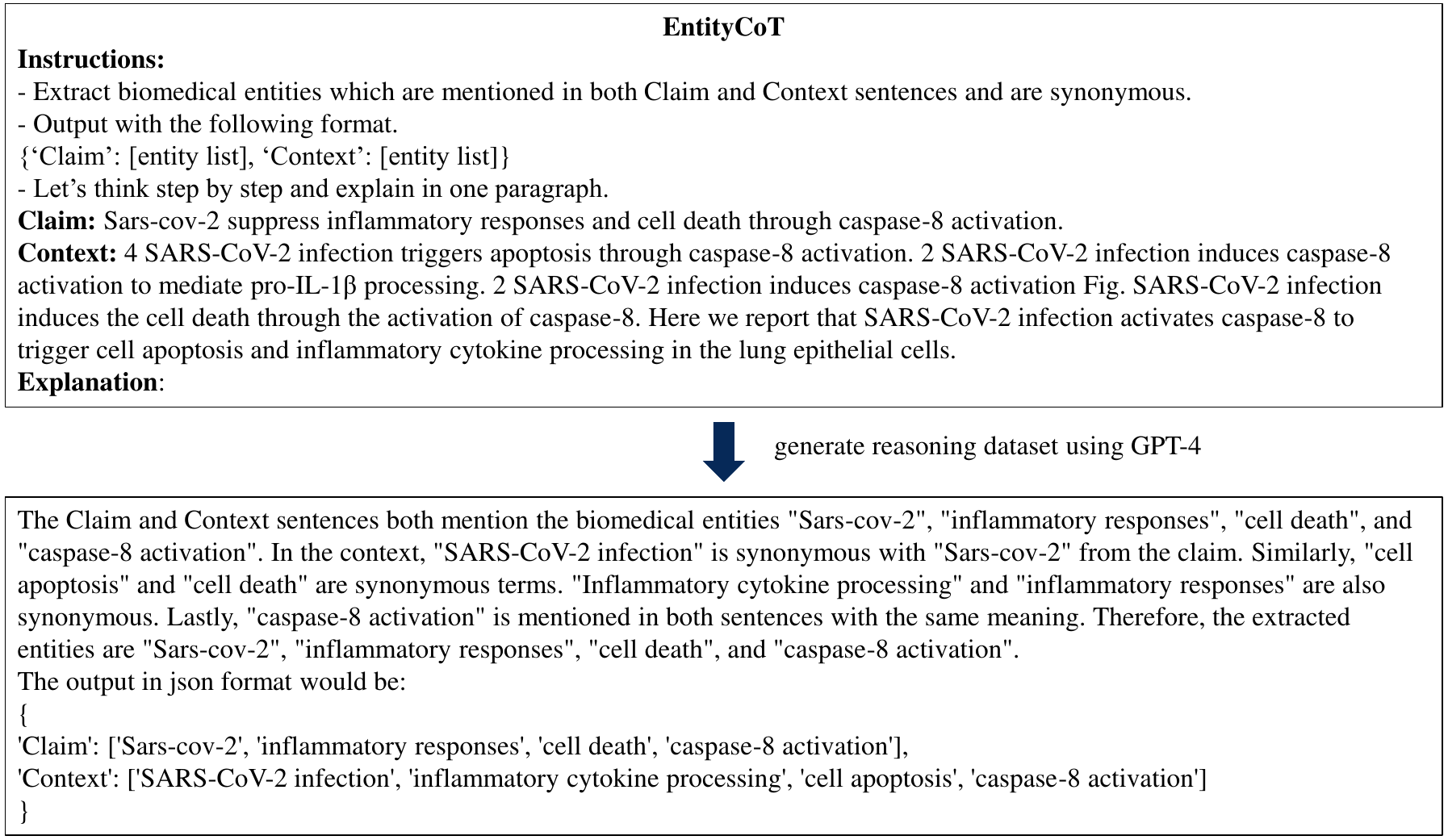}
    \caption{An example of EntityCoT dataset on the COVID-Fact.}
    \label{app_EntityCoT}
\end{figure*}

\begin{figure*}[]
    \centering
    \includegraphics[width=\textwidth]{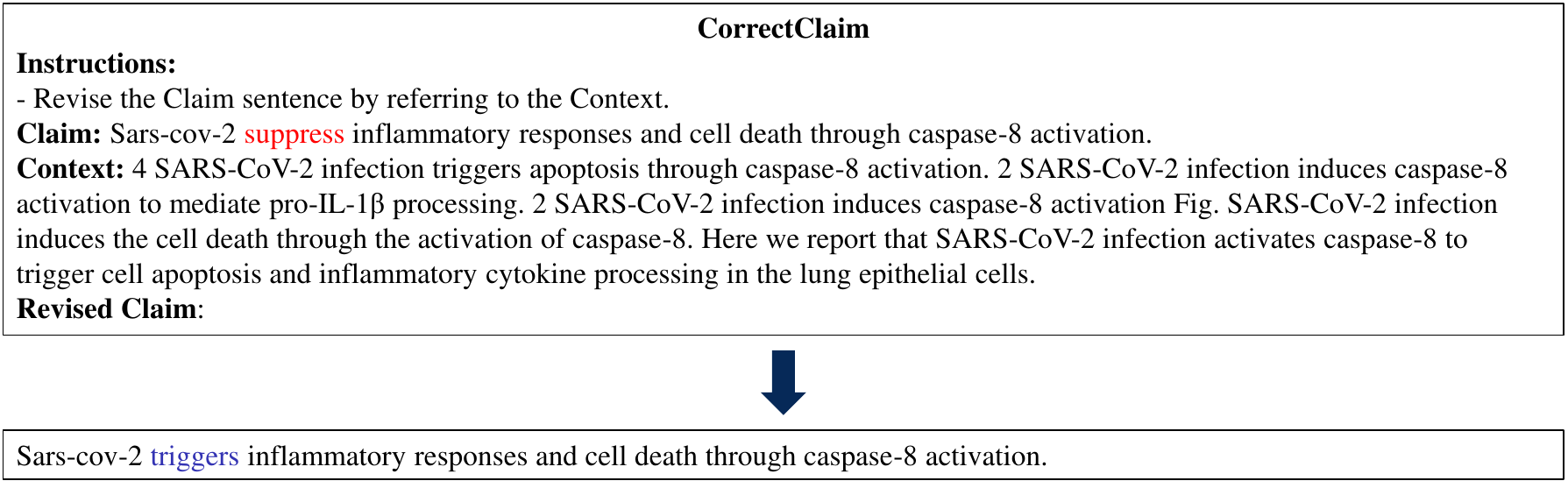}
    \caption{An example of CorrectClaim dataset on the COVID-Fact.}
    \label{app_CorrectClaim}
\end{figure*}

\begin{figure*}[]
    \centering
    \includegraphics[width=\textwidth]{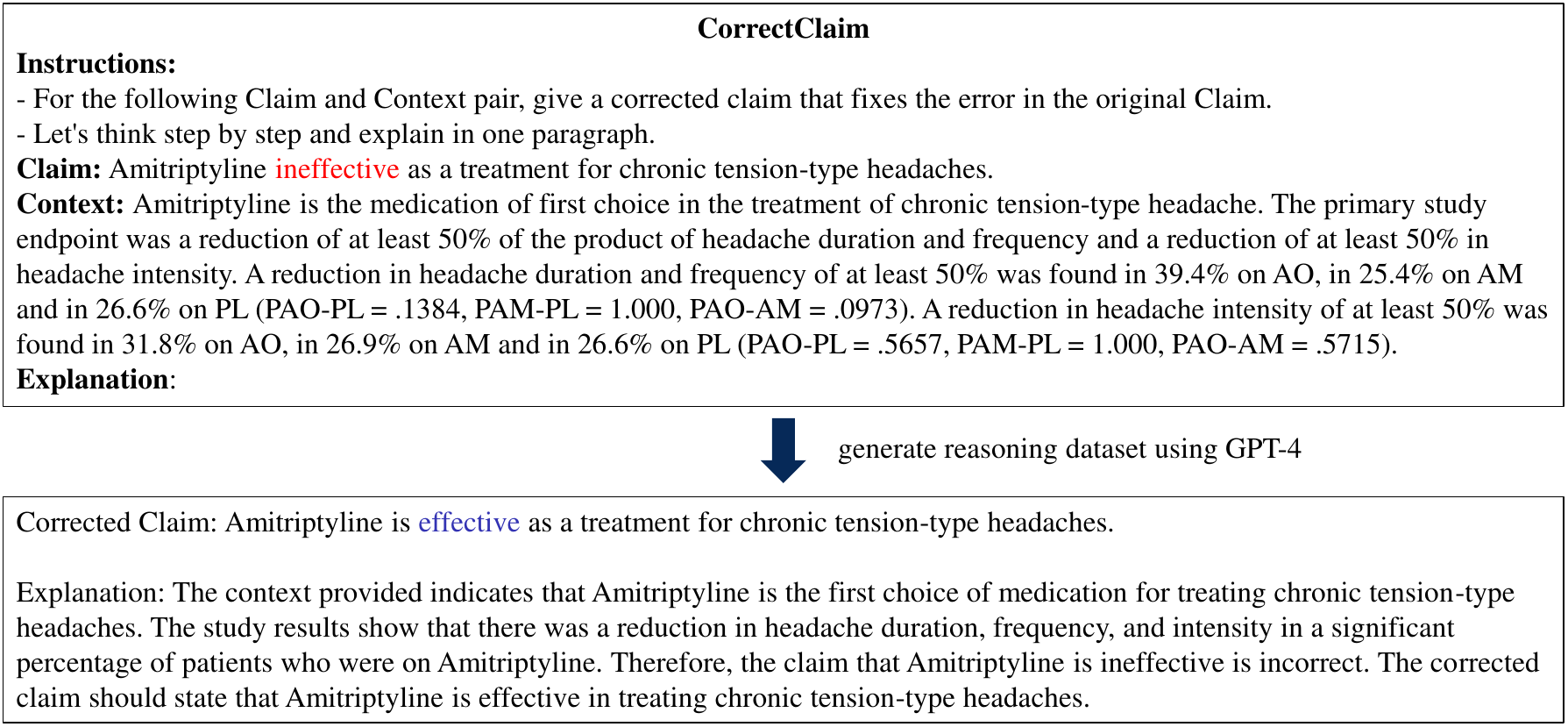}
    \caption{An example of CorrectClaim dataset using GPT-4 on the SciFact.}
    \label{app_CorrectClaim_gpt4}
\end{figure*}

We construct three reasoning datasets tailored for fact-checking: DifferenceCoT, EntityCoT, and CorrectClaim. Figure~\ref{app_DifferenceCoT}, ~\ref{app_EntityCoT}, and ~\ref{app_CorrectClaim} show the entire input and output of DifferenceCoT, EntityCoT, and CorrectClaim, respectively on the COVID-Fact dataset. 
For generating DifferenceCoT and EntityCoT, the input to GPT-4 encompasses task instruction, claim, and context, and the ground truth output is the GPT-4 result. 
For generating CorrectClaim, the input contains task instructions, claim, and context, and the ground truth output is the true claim for the given evidence. 
As SciFact dataset does not consistently offer both true and false claims for each evidence, we generate CorrectClaim using the GPT-4 API as shown in Figure~\ref{app_CorrectClaim_gpt4}.

\subsection{Examples of Fact-Checking Datasets}

\begin{figure*}[h!]
    \centering\small
    \includegraphics[width=\textwidth]{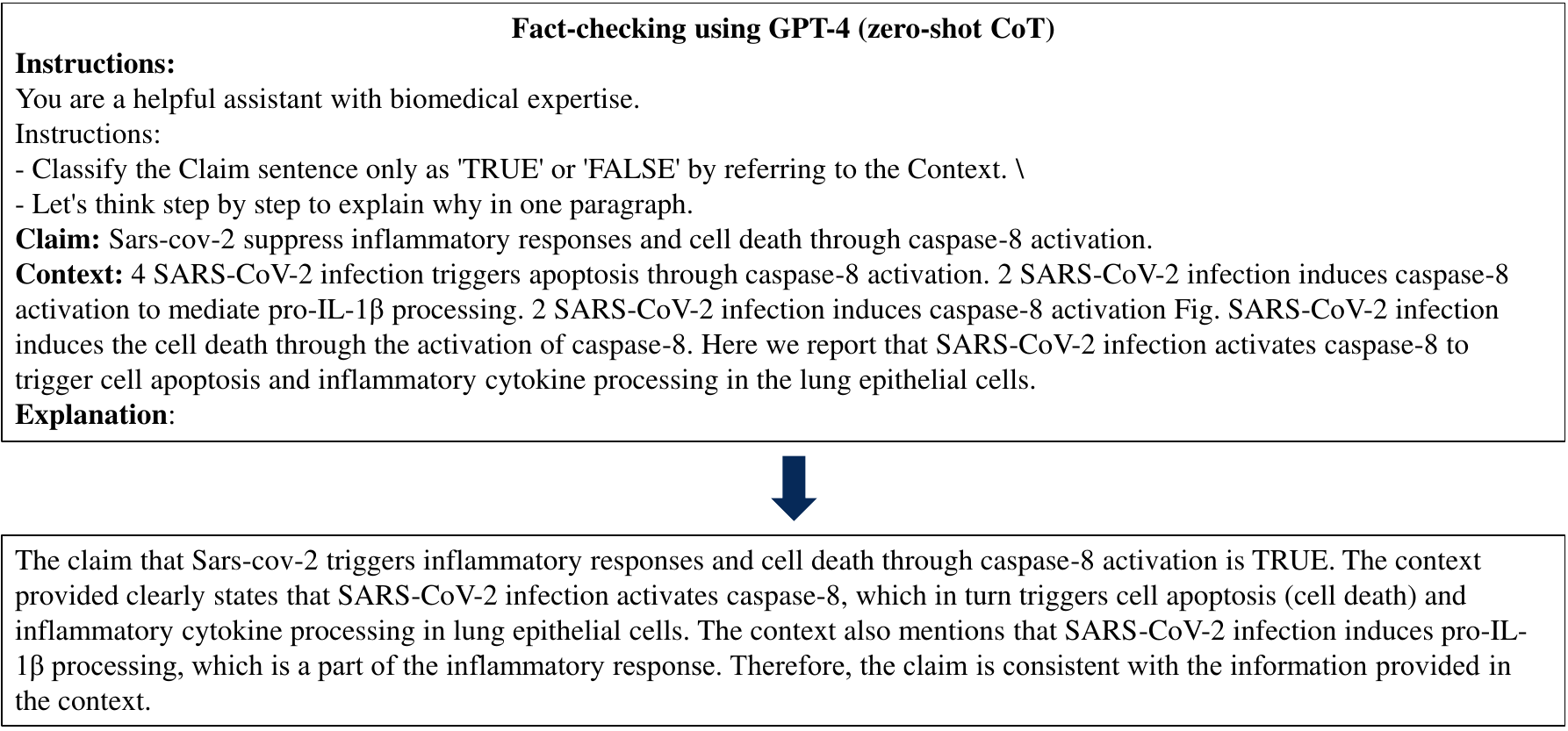}
    \caption{An example of fact-checking dataset when using zero-shot CoT with GPT-4.}
    \label{fc_zeroshot_prompt_GPT4}
\end{figure*}

\begin{figure*}[h!]
    \centering
    \includegraphics[scale=0.55]{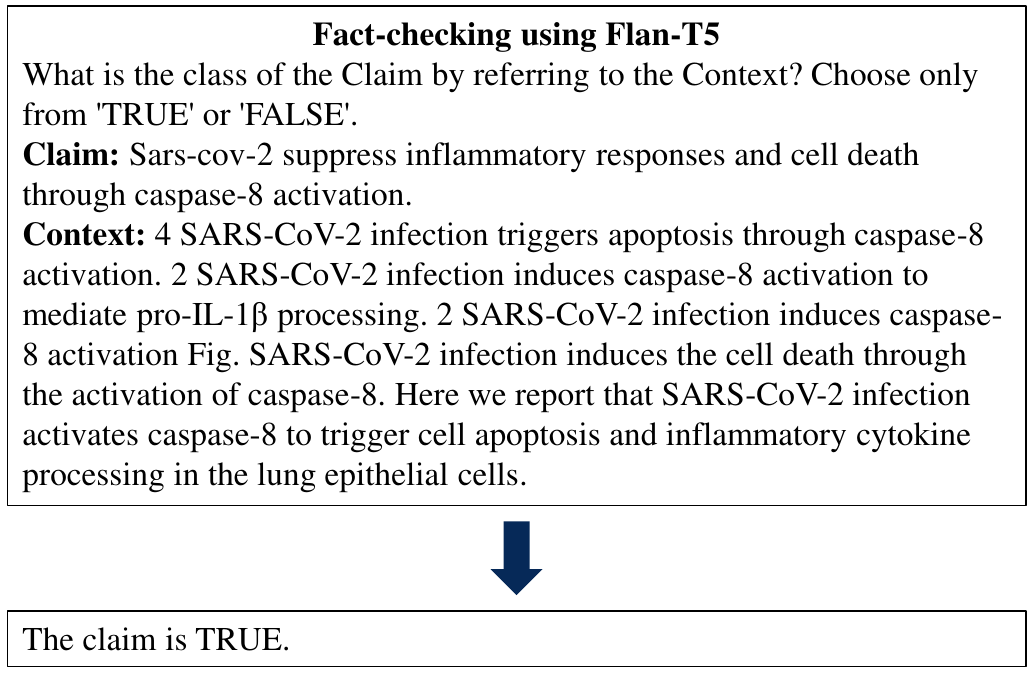}
    \caption{An example of fact-checking dataset for fine-tuning Flan-T5 models.}
    \label{fc_prompt_Flan-T5}
\end{figure*}

\begin{figure*}[h!]
    \centering
    \includegraphics[scale=0.55]{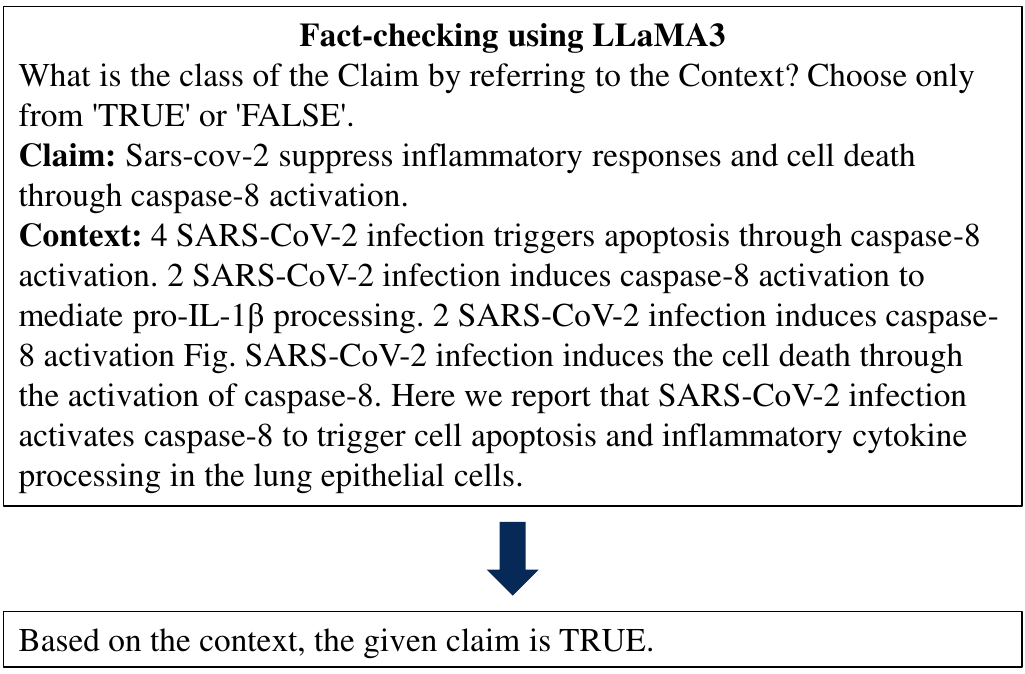}
    \caption{An example of fact-checking dataset for fine-tuning LLaMA3 models.}
    \label{fc_prompt_LLaMA3}
\end{figure*}

The input and output of the fact-checking dataset varies depending on the settings. 
An example of using zero-shot CoT with GPT-4 API is shown in Figure~\ref{fc_zeroshot_prompt_GPT4}. 
Figure~\ref{fc_prompt_Flan-T5} and Figure~\ref{fc_prompt_LLaMA3} show an example for fine-tuning the Flan-T5 model and LLaMA3 model, respectively.

\section{Experimental Settings}

\subsection{Fine-tuning Reasoning LoRAs}
All experiments use two RTX 3090 GPUs to fine-tune the three reasoning LoRAs with a fixed seed of 42 for reproducibility.
The LoRA and qLoRA configurations use 16 as the rank parameter, 32 as $\alpha$, and 0.05 as dropout for all models.
The batch size is 1, and the gradient accumulation step is 8. 

For Flan-T5 models, we employ early stopping with a patience of 3, selecting the epoch yielding the best ROUGE-Lsum score on the development set throughout the 20 training epochs.
The learning rate is $1e-3$, the warmup ratio is 0.1, and the weight decay is 0.01, and we adopt the adafactor optimizer coupled with a cosine scheduler. Depending on the dataset length, we set the maximum source and target length as 1200 and 512.

For the LLaMA3 model, we implement early stopping with a patience of 3, choosing the epoch that shows the lowest loss on the development set throughout 10 training epochs.
Using the ROUGE-Lsum score instead of loss leads to overfitting.
The learning rate is $2e-4$, the warmup ratio is 0.1, and the weight decay is 0.01, and we adopt the paged adamw 32bit optimizer alongside a cosine scheduler. Depending on the dataset length, we set the maximum length as 2048.

Table~\ref{lora_reasoning_time} presents the training time and inference time for each test instance.
Although Flan-T5-xxl shows the best performance on COVID-Fact, its extensive training and inference times make LLaMA3-8B a more practical choice.
Therefore, we experiment with LLaMA3-8B for SciFact.

\begin{table}[h]
\centering
\resizebox{\columnwidth}{!}{%
\begin{tabular}{cccc}
\hline
\textbf{Dataset} & \textbf{Reasoning LoRA} & \textbf{Model} &  \textbf{Train/Test Time} \\
\hline
\multirow{9}{*}{\begin{tabular}[c]{@{}c@{}}COVID- \\Fact\end{tabular}} & 
\multirow{3}{*}{DifferenceCoT}
& Flan-T5-large & 6h 50m / 7s \\ \cline{3-4}
& & Flan-T5-xxl & 35h 50m / 21s \\ \cline{3-4}
& & LLaMA3(8B) & 2h 27m / 12s \\ \cline{2-4}
& \multirow{3}{*}{EntityCoT}
& Flan-T5-large & 8h 8m / 7s \\ \cline{3-4}
& & Flan-T5-xxl & 18h 35m / 4m 21s \\ \cline{3-4}
& & LLaMA3(8B) & 2h 37m / 12s \\ \cline{2-4}
& \multirow{3}{*}{CorrectClaim}
& Flan-T5-large & 3h 46m / 1s \\ \cline{3-4}
& & Flan-T5-xxl & 16h 9m / 4s \\ \cline{3-4}
& & LLaMA3(8B) & 1h 1m / 3s \\
\hline
\multirow{3}{*}{SciFact}
& DifferenceCoT & LLaMA3(8B) & 59m / 25s \\ \cline{2-4}
& EntityCoT & LLaMA3(8B) & 1h / 22s \\ \cline{2-4}
& CorrectClaim & LLaMA3(8B) & 53m / 21s \\
\hline
\end{tabular}%
}
\caption{Training and inference time comparison for reasoning LoRAs.}
\label{lora_reasoning_time}
\end{table}

\subsection{Fine-tuning on Fact-Checking}

All experiments use two RTX 3090 GPUs to fine-tune fact-checking with a fixed seed of 42 for reproducibility.
The LoRA and qLoRA configurations use 16 as the rank parameter, 32 as $\alpha$, and 0.05 as dropout for all models.
For all models, we employ early stopping with a patience of 3, selecting the epoch yielding the best macro-f1 score on the development set.

For the Flan-T5-large model, the learning rate is $1e-3$, the warmup ratio is 0.1, and the weight decay is 0.01. The training epoch is 20, and we adopt the adafactor optimizer coupled with a cosine scheduler. 
The batch size per device is 1, and the gradient accumulation step is 4. 
We set the maximum source and target length as 1200 and 512. 

For the Flan-T5-xxl model, the learning rate is $3e-4$, the warmup ratio is 0.1, and the weight decay is 0.01. The training epoch is 10, and we adopt the adafactor optimizer coupled with a cosine scheduler. 
The batch size per device is 1, and the gradient accumulation step is 8. 
We set the maximum source and target length as 1200 and 512.

For the LLaMA3 model, the learning rate is $2e-4$, the warmup ratio is 0.1, and the weight decay is 0.01. The training epoch is 10, and we adopt the paged adamw 32bit optimizer along with a cosine scheduler. 
The batch size per device is 1, and the gradient accumulation step is 8. 
We set the maximum length as 1024.

\section{Case Study}
\begin{sidewaysfigure}
    \centering
    \includegraphics[width=\textwidth]{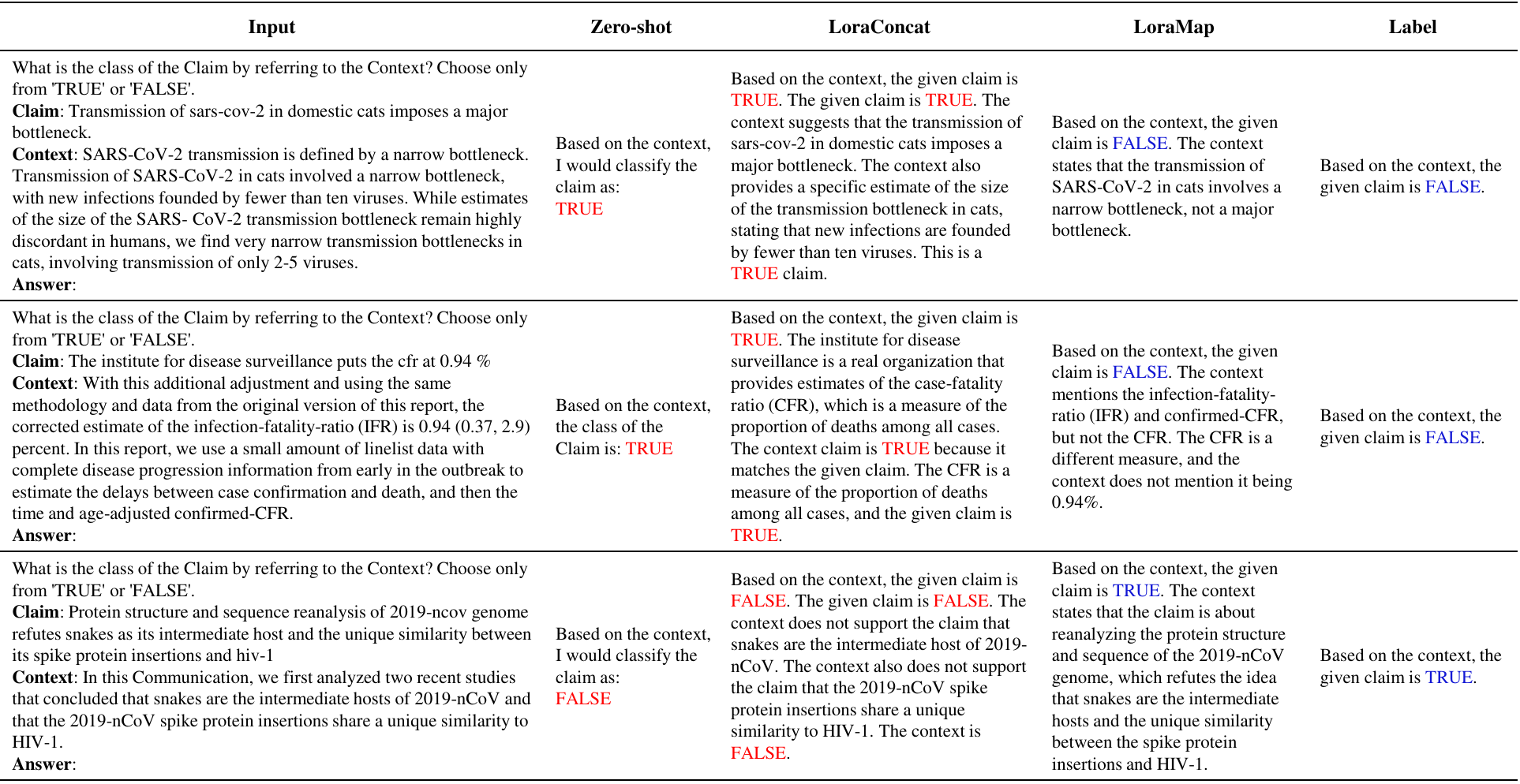}
    \caption{Examples of outputs on the COVID-Fact dataset.}
    \label{app_case_study_covidfact_llama}
\end{sidewaysfigure}

\begin{sidewaysfigure}
    \centering
    \includegraphics[width=\textwidth]{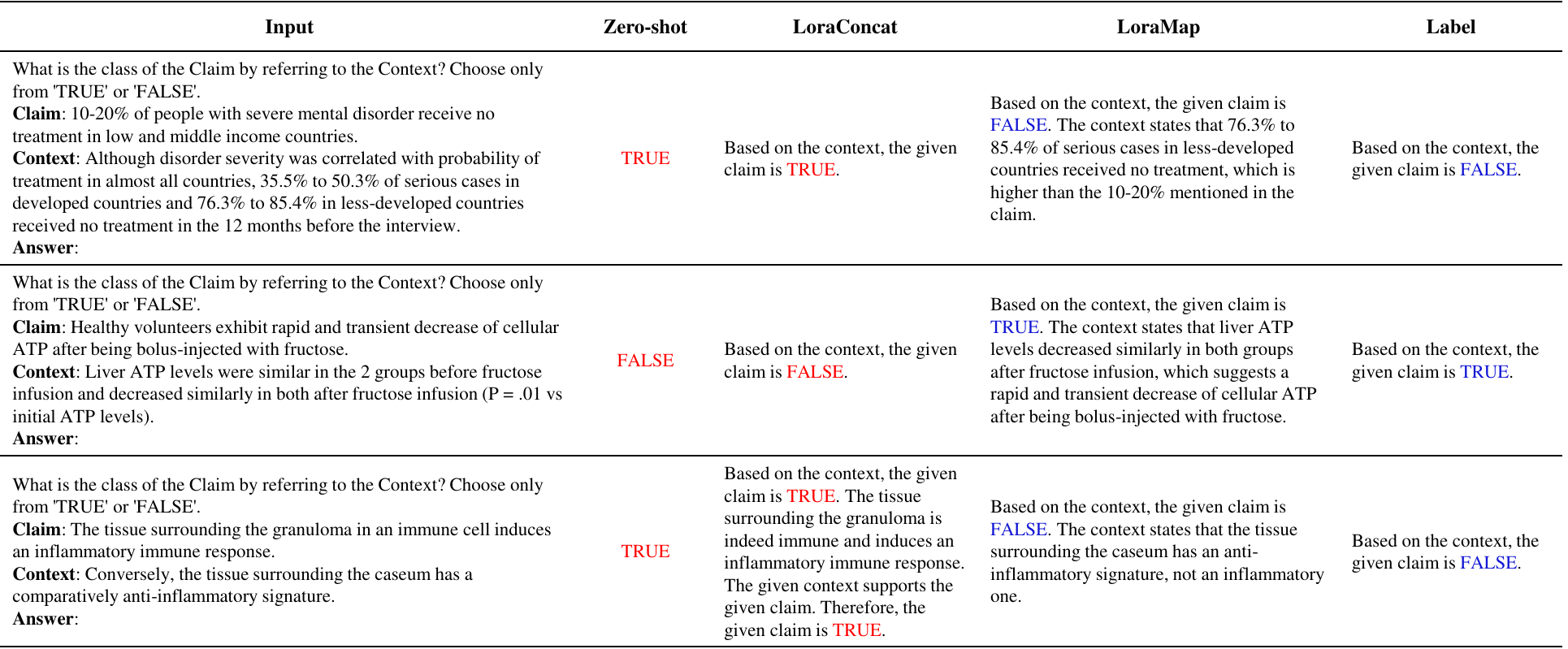}
    \caption{Examples of outputs on the SciFact dataset.}
    \label{app_case_study_scifact_llama}
\end{sidewaysfigure}

The cases where the output of LoraMap is correct, but zero-shot and LoraConcat are incorrect on the COVID-Fact and SciFact datasets are shown in Figure\ref{app_case_study_covidfact_llama} and \ref{app_case_study_scifact_llama}, respectively. While zero-shot only performs classification on the claim, LoraConcat and LoraMap tend to provide explanations, possibly due to the use of reasoning LoRAs. LoraConcat often attempts to incorrectly classify while explaining. It is quite surprising that LoraMap attempts to generate explanations, even though the trained label does not include an explanation part. In the future, it may be beneficial to use labels with explanations to improve model explainability.

\onecolumn

\clearpage

\twocolumn

\end{document}